\documentclass[conference]{IEEEtran}
\IEEEoverridecommandlockouts
\usepackage{cite}
\usepackage{amsmath,amssymb,amsfonts}
\usepackage{algorithmic}
\usepackage{algorithm}
\usepackage{graphicx}
\usepackage{textcomp}
\usepackage{xcolor}
\usepackage{multirow}
\def\BibTeX{{\rm B\kern-.05em{\sc i\kern-.025em b}\kern-.08em
    T\kern-.1667em\lower.7ex\hbox{E}\kern-.125emX}}
\begin{document}

\title{Event Detection in Noisy Streaming Data with Combination of Corroborative and Probabilistic Sources\\
\thanks{}
}

\author{\IEEEauthorblockN{Abhijit Suprem}
\IEEEauthorblockA{\textit{School of Computer Science} \\
\textit{Georgia Institute of Technology}\\
Atlanta, USA\\
asuprem@gatech.edu}
\and
\IEEEauthorblockN{Calton Pu}
\IEEEauthorblockA{\textit{School of Computer Science} \\
\textit{Georgia Institute of Technology}\\
Atlanta, USA \\
calton.pu@cc.gatech.edu}
}

\maketitle

\begin{abstract}
Global physical event detection has traditionally relied on dense coverage of physical sensors around the world; while this is an expensive undertaking, there have not been alternatives until recently. The ubiquity of social networks and human sensors in the field provides a tremendous amount of real-time, live data about true physical events from around the world. However, while such human sensor data have been exploited for \textit{retrospective} large-scale event detection, such as hurricanes or earthquakes, they has been limited to no success in exploiting this rich resource for general physical event detection. 

Prior implementation approaches have suffered from the concept drift phenomenon, where real-world data exhibits constant, unknown, unbounded changes in its data distribution, making static machine learning models ineffective in the long term. We propose and implement an end-to-end collaborative drift adaptive system that integrates  \textit{corroborative} and probabilistic sources to deliver real-time predictions. Furthermore, out system is adaptive to concept drift and performs automated continuous learning to maintain high performance. We demonstrate our approach in a real-time demo available online for landslide disaster detection, with extensibility to other real-world physical events such as flooding, wildfires, hurricanes, and earthquakes.
\end{abstract}


\section{Introduction}
Physical event detection, such as extreme weather events or traffic accidents have long been the domain of static event processors operating on numeric sensor data or human actors manually identifying event types. However, the emergence of big data and associated data processing and analytics tools and systems have led to several applications in large-scale event and trend detection in the streaming domain \cite{dis_mgmt_hagen,dis_mgmt_imran,dis_mgmt_nagy,dis_mgmt_sakaki,disease_mgmt_hirose,bursty_dis_mgmt,flood_detection}. However, it is important to note that many of these works are a form of retrospective analysis, as opposed to true \textit{real-time} event detection, since they perform analyses on cleaned and processed data within a short-time frame in the past, with the assumption that their approaches are sustainable and will continue to function over time. 

This is an unrealistic assumption due to the concept drift phenomenon, where real-world data exhibits continuous changes in its distribution. The concept drift phenomenon has been well documented \cite{conc_drift_active_shan,conc_drift_almeida,conc_drift_costa,conc_drift_demello,conc_drift_windows,gama_drift_a,gama_drift_c,gft_fail_b,gft_fail_a,gft_fail_c}. In effect, changes in data distribution render machine learning algorithms obsolete over time and classification models require constant fine-tuning for effective performance. As such, existing big data analytics have focused on larger scale events or trend analysis where learning models can be updated with human feedback. 

As such, most applications for streaming data rely on non-adversarial assumptions about their data content:
\begin{itemize}
	\item the streaming data is of high quality, with little to no noise; in this case, human labeling is easier and weak-supervision~\cite{weak_supervision} using trend analysis or statistical distributions can be exploited to create new labeled data
	\item the concept drift direction, type, and scale are known; in effect, some approaches presuppose knowledge of dataset shift, which is not a realistic real-world assumption
	\item there is immediate and proportional feedback available to perform model correction; the streaming domain's data volume is too large to enable for proportional feedback
	\item the streaming data exhibits strong-signal characteristics, where the desired event's signals (or features) are well separated from irrelevant signals; in our case, we focus on the weak-signal case where the relevant data is dwarfed by irrelevant data and noise.
\end{itemize}

We present an system for adapting to real-world evolving data that uses a combination of \textit{corroborative sources} and \textit{probabilistic supporting sources} to perform real-time event detection that avoids deterioration under noisy, drifting conditions. We demonstrate our system in a case study with disaster detection as the physical event of choice; our system is able to detect events under a variety of categories, such as landslides, floodings, wildfires, and earthquakes in real-time. Additionally, our system LITMUS\footnote{A demo is available at: https://grait-dm.gatech.edu/demo-multi-source-integration/}, is drift adaptive and continuously updates itself against adversarial drift without human intervention. Specifically, we address the closed-dataset assumptions described:
\begin{itemize}
	\item we rely on low quality streaming data from social networks such as Twitter and Facebook, which consist primarily of noisy short-text streams \cite{short_text}\cite{short_text_sriram} with large amounts of misinformation and disinformation
	\item we do not pre-suppose drift; instead we assume unknown and unbounded concept drift due in part to lexical diffusion \cite{lex_diff} and random shifts in user behavior
	\item we do not rely on human feedback for our system's learning model updates due to its infeasibility in the streaming domain; manually labeling of even 0.01\% of streaming web data from Twitter ($>$500M samples per day) will require more than 20 workers to work continuously for 8 hours each day
	
	\begin{figure*}[t]
		\centering
		\includegraphics[width=0.8\linewidth]{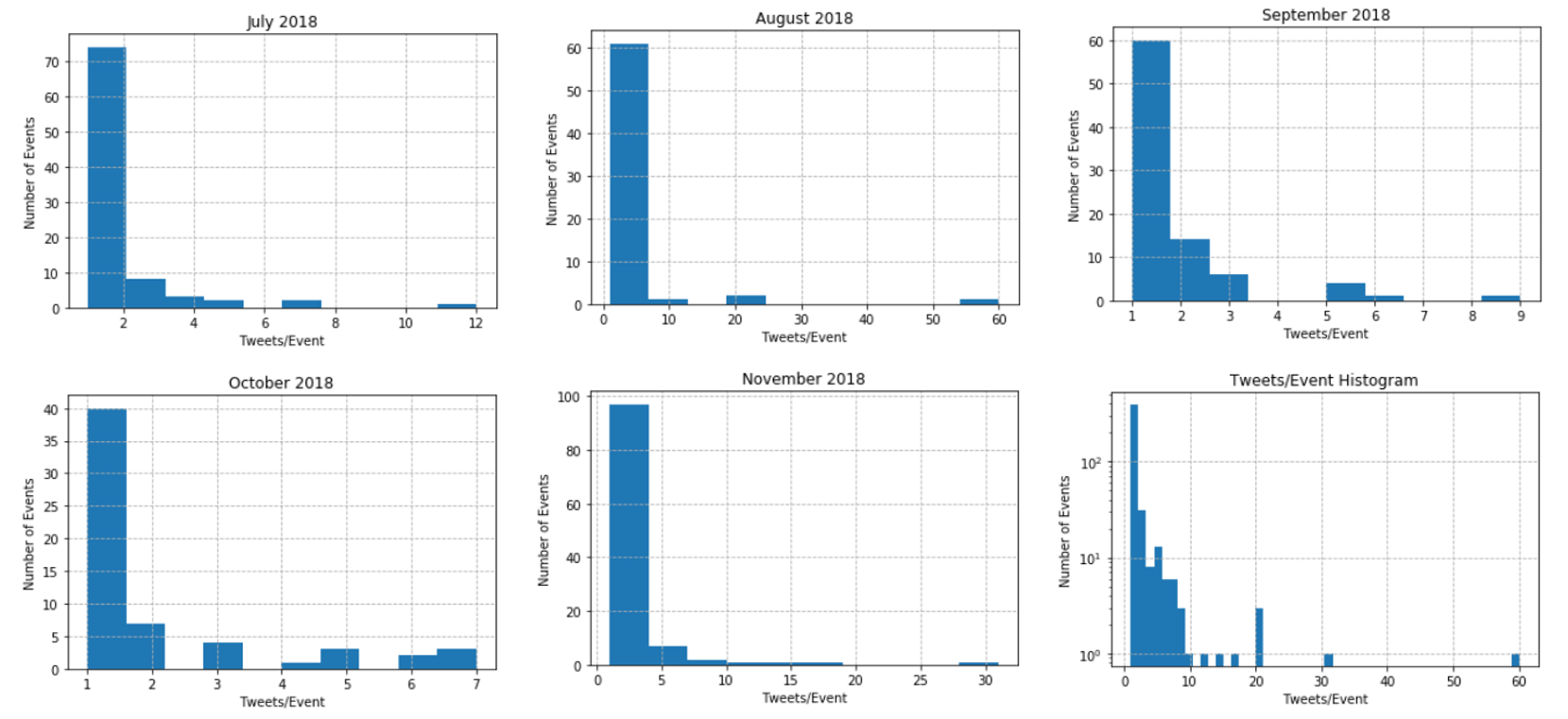}
		\caption{Most events have 1 post associated with them. More than 95\% of event are detected with less than 10 posts per event. Event detection confidence is built over time as more posts are discovered.}
		\label{fig:tweethisto}
	\end{figure*}

	\item we demonstrate efficacy with weak-signal events with an abundance of irrelevant data and noise - our disaster dataset is an ongoing collection of live social and news feeds, and even with keyword search and filtering on disaster type, almost 94\% of data are drifting noise with time-varying characteristics that must be eliminated with fast-updating learning models
\end{itemize}

We make the distinction between \textit{weak-signal} and \textit{strong-signal} events as follows: strong-signal events have signals (or features) that are easily separable; for example, the earthquake detection approach in\cite{dis_mgmt_sakaki} relies on the fact that each earthquake is followed by several hundreds or thousands of tweets. Similarly, Google Flu Trends, another example of event detection that deteriorated due to concept drift ~\cite{assed,gft_fail_a,gft_fail_b,gft_fail_c}, focused on flu detection by matching search terms across the entire United States. We focus on weak-signal events such as landslides, floodings, and wildfires: these events have the same real-world impact int terms of damages and costs; however they are numerous and each instance of an event is lost in streaming noise. We show in Figure~\ref{fig:tweethisto} the relation between our event detection to the number of tweets per event: most events are associated with a single tweet or two tweets - a far cry from the hundreds or millions of social sensors used by ~\cite{dis_mgmt_sakaki} and \cite{gft_fail_c}. We perform detection in these noisy and drifting conditions, where our approach outperforms static models by over 350\% in event detection under drifting conditions.

We present the following contributions:
\begin{itemize}
	\item We propose a system for end-to-end event detection using a combination of \textit{corroborative sources} and \textit{probabilistic supporting sources}
	\item We implement a collaborative teamed-classifier approach for physical event detection that performs continuous learning and adaptation without human intervention. Our approach is able to detect concept drift and perform the appropriate training data generation, labeling, and model fine-tuning to prevent classifier deterioration without any bottleneck from human labelers or fine-tuners
	\item We demonstrate the efficacy of our system on weak-signal events with  significant amounts of noise and concept drift. A demo is available on LITMUS at $$\text{https://grait-dm.gatech.edu/demo-multi-source-integration/}$$
\end{itemize}

\section{Related Work}

\subsection{Concept Drift}
Recent approaches for drift adaptation usually use synthetic data to validate procedures \cite{conc_drift_active_shan,conc_drift_almeida,conc_drift_costa,conc_drift_demello}. Synthetically generated data is perturbed to include specific, known forms of drift, such as gradual, cyclic, or sudden drift. Under these constraints, there exist several mechanisms for concept drift adaptation with physical sensors containing numeric data. 

\textbf{Windowing} is a common technique for adaptation that uses multiple sliding windows over time. This approach uses several data memories, or windows of different lengths over an incoming data stream; each window has its own classifier. The \textbf{SAM-KNN} algorithm uses nearest neighbor approach to select the window closest to a new data sample for classification \cite{samknn}. Nested windows are used in \cite{conc_drift_windows} to obtain multiple training sets over the same data that each exclude a region of the data space. 
\textbf{Adaptive Random Forests} augment the traditional random forest classifier with a built-in explicit drift detector (requiring labels). Drift detection leads to forest pruning in the ensemble to remove tress that have poor performance on the drifted data. The pruned forest is subsequently updated with new weak classifiers to complete the ensemble \cite{arf}. 
The \textbf{Knowledge Maximized Ensemble} (KME) uses both off-the-shelf and their own drift detectors to recognize multiple forms of drift simultaneously. Models are updated when enough training data is collected and removed if they perform poorly on subsequent drifted data \cite{kme}.
Most methods approach concept drift with an eye towards detection and subsequent normalization. Updating or rebuilding a machine learning model facing drift involves two bottlenecks in the classification pipeline: data labeling and model training; data labeling is the greater challenge due to its oracle requirements. Such wait-and-see models that perform corrections once errors have been detected entail periodic performance degradation before they are corrected with model updates; this may be infeasible in mission-critical applications. Active learning strategies counteract this bottleneck in part \cite{conc_drift_active_shan}; the trade-off is between highly accurate models and clustered, knowledge-agnostic representations that consider data on distance without subject matter expertise. 

\subsection{Physical Event Detection}

Earthquake detection using social sensors was initially proposed in \cite{dis_mgmt_sakaki}. There have also been attempts to develop physical event detectors for other types of disasters, including flooding \cite{flood_detection}, flu \cite{gft_fail_a, gft_fail_b, gft_fail_c}, infectious diseases \cite{dis_mgmt_hagen}, and landslides \cite{litmus_a}. In most cases, the works focus on large-scale disasters or health crises, such as earthquakes, hurricanes \cite{dis_mgmt_thom}, and influenza that can be easily verified and have abundant reputable data.. Our application is general purpose, as it can handle small-scale disaster such as landslides and large-scale disasters. The existing approaches also assume data without concept drift. However such assumptions, made in Google Flu Trends (GFT) \cite{gft_fail_a,gft_fail_b,gft_fail_c} degrade in the long term. GFT was originally created to complement the CDC’s flu tracking efforts by identifying seasonal trends in the flu season ~\cite{gft_fail_c}. Failure to account for seasonal changes in event characteristics led to increasing errors over the years, and by 2013, GFT missed the trends by 140\%. This error has been attributed to exclusion of new data from CDC, changes in the underlying search data distribution itself, and cyclical data artifacts ~\cite{gft_fail_c,gft_fail_a, gft_fail_b}.

\section{Data}
Our system uses a combination of \textit{corroborative sources} and \textit{probabilistic supporting sources}. We first make the distinction between the two before describing our data collection process.

\paragraph{Corroborative source}
We define a corroborative sources as a dedicated physical or web sensors that provides annotated physical event information that can be crawled or scraped. Such physical sensor data is often  structured, e.g. government agency reports about disasters. Web-based corroborative sources include news articles which are often tagged with keywords and due to their fact-based nature, inherently included misinformation checking. However, corroborative source latency in information availability makes them unsuited for real-time physical event detection; since corroborative sources provide event conformation after their own corroboration, there are delays in information dissemination. Such sources also do not have global or dense coverage due to funding limits. 
\paragraph{Probabilistic supporting source}
We consider any source without corroboration a probabilistic supporting source due to the inherent uncertainty. These correspond to classifier predictions in a traditional ML environment. In our approach, we use an array of probabilistic supporting sources to more confidently predict events in the absence of corroborative sources. Additionally, our systems monitors these probabilistic supporting sources continuously for performance deterioration due to drift, and performs classifier updates and fine-tuning using data from corroborative sources. As such, we call this combination of two types of sources the \textbf{teamed-classifier} approach.

\subsection{Corroborative Sources}
In our case study of disaster detection, our corroborative sources are extensible based on the domain. We use a combination of physical and web sensors as our corroborative sources. 

As an example of corroborative source latency and limited coverage, the LITMUS system previously relied primarily on USGS landslide reports~\cite{litmus_a}. Since USGS no longer provides any landslide reports for such disasters, thee LITMUS system must compensate with other corroborative sources such as rainfall and earthquake data from USGS. We also use NOAA landslide predictions that are provided in high rainfall regions. Since these physical sensors do not have dense, global coverage, we also use web-based corroborative sources such as news articles crawled from aggregators (Google News and Bing News APIs). We adapt the news streaming and processing approach from \cite{assed} for data collection.

\subsection{Probabilistic Supporting Sources}
Our probabilistic supporting sources are a group of temporally evolving machine learning classifiers trained to classify events from short-text streams from Twitter, Facebook, and other social networks. In contrast to an ensemble approach, we only use the classifiers that are most effective on a data item; due to concept drift, we keep a history of classifiers at different points of training over time as well as each classifier's performance to better identify high quality classifiers at any time. The raw social sensor data is streamed from Twitter and Facebook, with web crawlers leveraged for the latter to improve retrieval efficiency. LITMUS performs metadata processing using the streaming and extraction approach in \cite{assed}. We note that even with keyword filtering (e.g. \textit{landslide}, \textit{mudslide}, \textit{rockslide} for landslides, and \textit{flood} and \textit{rain} for flooding), over 90\% of the streamed data are not relevant to our desired disaster events. In each case, there is linguistic noise that hides the true events:
\begin{itemize}
	\item \textit{Landslide} refers to both the disaster event and election events. Additionally, there is a song with the same name by the \textit{Fleetwood Mac} band.
	\item \textit{Mudslide} refers to both the disaster event and an alcoholic cream drink
	\item \textit{Flood} is used to describe flooding events in conjunction with the more idiomatic usage
	\item \textit{Rain} similarly is used for heavy rain events, light rain events, and idiomatic usage, the latter of which is more common (e.g. \textit{raining on their parade})
\end{itemize}

\begin{figure}[h]
	\centering
	\includegraphics[width=\linewidth]{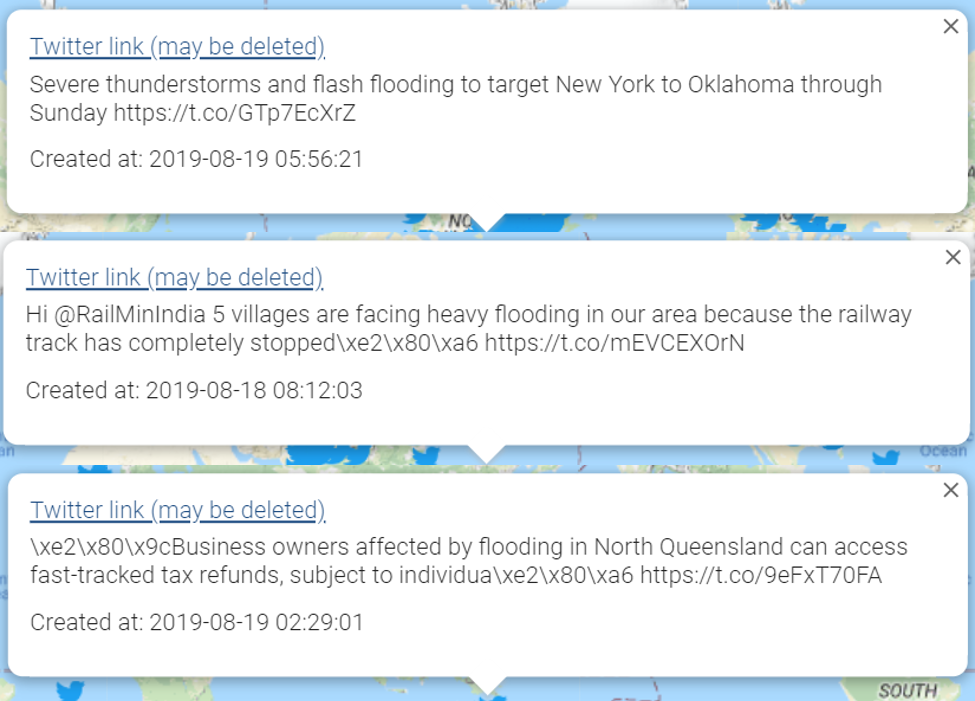}
	\caption{Flooding events detected by LITMUS}
	\label{fig:floodingtweet}
\end{figure}

\begin{figure}[h]
	\centering
	\includegraphics[width=\linewidth]{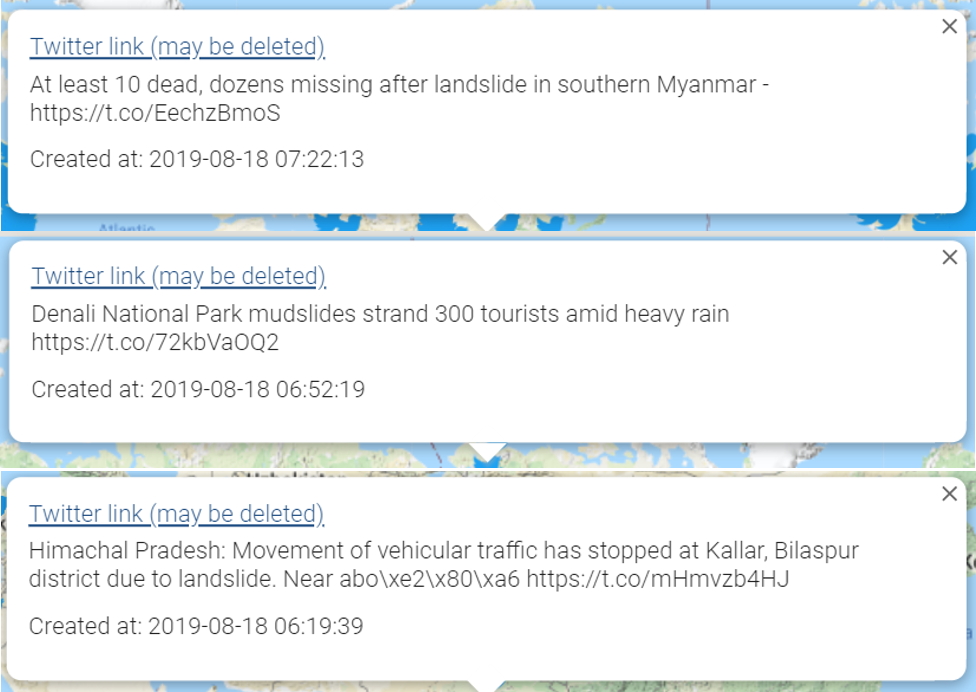}
	\caption{Landslide events detected by LITMUS}
	\label{fig:landslidetweet}
\end{figure}

\begin{figure}[h]
	\centering
	\includegraphics[width=\linewidth]{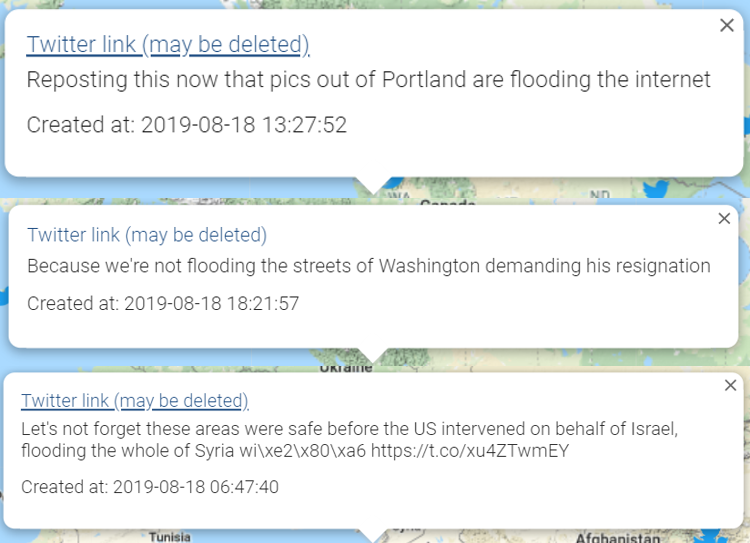}
	\caption{False positive event detection due to linguistic noise. Over time, LITMUS learns to adapt to such noise.}
	\label{fig:badtweet}
\end{figure}

We show some examples of correctly detected events from streaming web data in Figure~\ref{fig:floodingtweet} and ~\ref{fig:landslidetweet}. We also show examples of false positives due to linguistic noise in Figure~\ref{fig:badtweet}. Additionally, as we showed in Figure~\ref{fig:tweethisto}, most events have only one or two tweets or Facebook posts associated with them, requiring stronger detection capabilities for real-time event detection than retrospective trend-detection approaches in\cite{dis_mgmt_sakaki}.

\section{System Architecture}
We first describe our general system overview. We will then cover technical details about the integration of corroborative and probabilistic sources, as well as the unsupervised drift detection and adaptation algorithms we use. Finally, we'll cover our system implementation.

\subsection{System Overview}
\begin{figure}[h]
	\centering
	\includegraphics[width=\linewidth]{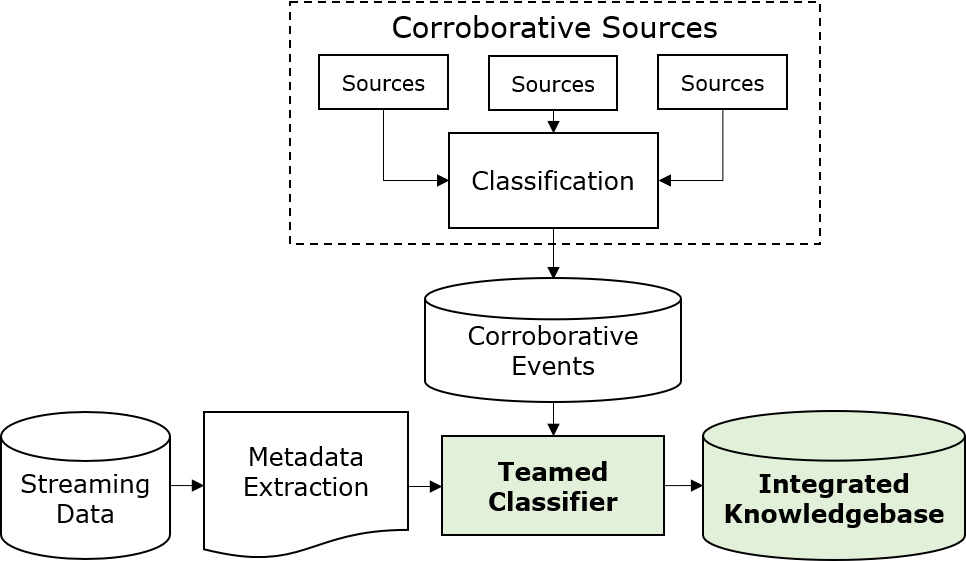}
	\caption{General system overview for corroborative and probabilistic source integration. We perform classification on corroborative sources using their own annotations and store them in a Corroborative Events database. Streaming Data is classified into relevant and irrelevant classes using teamed classifiers that combine corroborative events and probabilistic supporting sources operating on the streaming data. The detected events are stored in the Integrated Knowledgebase.}
	\label{fig:systemoverview}
\end{figure}

\begin{figure}[h]
	\centering
	\includegraphics[width=\linewidth]{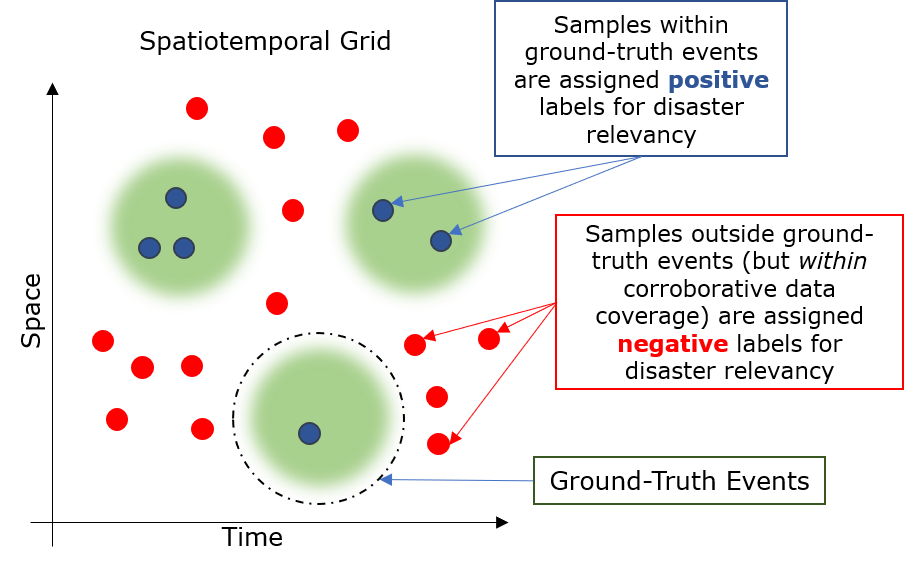}
	\caption{We can automatically label some of the streaming data points using corroborative events: streaeming data points (small circles) in the same spatio-temporal coordinates as a corroborative event (green zones) get the same label as the corroborative event. For example, if we get corroboration of a landslide event in Austin, Texas, then any tweet or Facebook post during the same time and location mentioning landslides is more likely about the disaster event as opposed to election landslides or the song \textit{Landslide}.}
	\label{fig:spatiotemporal}
\end{figure}

We show a general system overview in Figure~\ref{fig:systemoverview}. The \textbf{teamed classifier} is dynamically constructed weighted ensemble of classifiers that are most relevant for a given data point. Classifiers consist of machine learning models trained on subsets of streaming data collected since system inception, as well as spatio-temporal filters based on corroborative events. The latter uses the following intuition: if an event is detected from corroborative sources, then any streaming data point that exists in the same spacio-temporal coordinates as the corroborative event can be automatically labeled as a relevant, or conversely, an irrelevant data point (see Figure~\ref{fig:spatiotemporal}).

\subsubsection{Corroborative Classification}
We perform corroborative classification using annotations provided by corroborative sources themselves. As an example, NOAA provides landslide predictions in high rainfall regions. We take high probability predictions as ground truth for landslide events in the future, since NOAA predictions are high confidence sources (as defined in \cite{assed}) and can corroborate events detected in streaming data. Similarly, NOAA rainfall data covers flooding during extreme climate events. We use these annotations as ground truth corroborative events.

\subsubsection{Streaming Data Classification}
We use an array of machine learning models for streaming data classification. Given a short-text stream from the raw stream, we select the top $k$ most relevant classifiers from set of all classifiers stored in LITMUS; we empirically set $k=5$. Each classifier is weighted based on its relevancy to the data point (we cover this weighting scheme in the next section) and the dynamically created ensemble is used for the streaming data point classification.

\subsection{Teamed Classifier Selection}
We determine relevancy of a classifier to a data point using its performance on similar data points. This requires two steps: (i) drift detection to identify changes in the data distribution, and therefore, distance between data points, and (ii) classifier generation and selection in case of drift detection. In the second step, we perform classifier generation if the drift has not been seen, e.g. gradual or flash drift as described in \cite{gama_drift_a}. We perform classifier selection if the drift has been seen before, as in the case of cyclic or periodic drift \cite{gama_drift_a}.

\subsubsection{Drift detection}
We considered recent works on novelty detection or out-of-distribution detection \cite{oodd1,oodd2,oodd3,oodd4}. In our weak-signal focus, such approaches are not suitable where most of the  samples are noise. We also need to address virtual concept drift, where the distribution of both relevant and irrelevant points changes without changing the decision boundary itself. Under virtual drift, it is sufficient to fine tune a classifier instead of rebuilding it, which is a more expensive step. Since concept drift affects the underlying data distribution, our drift detection approach uses the Kullback-Leibler divergence test on two distribution windows - the set of data points a classifier is trained on and the current streaming window of incoming data points. Comparison of the two distributions yields the distribution divergence metric, which we use as distance between a data point and a classifier. We perform the comparison on a high-density band of points for each window, with the band defined as follows: let $D'$ be the points ${x_1,x_2,...,x_i}$ in window $w'$, with a mean (or centroid) $\bar{D'}=N^{-1}\sum_N x_i$

Then, let $f_D(x)$ be the continuous density function of any $D$, where we estimate it on the distribution of distances of any point in $D$ from its centroid $\bar{D}$ normalized to $[0,1]$. Then, the $\Delta$-density band of $D$, with $\Delta\in[0,1]$, is a band around the centroid that contains $\Delta$ probability mass of the data window; e.g. if $\Delta=0.6$, the the $\Delta$-band contains 60\% of the points in $D$. We consider this as a banded region $[\delta_l, \delta_h]$, where $0\leq\delta_l < \delta_h\leq 1$, and calculate the region bounds as:

\begin{equation}
\int_{\delta_l}^{\delta_h}f_D(x)dx=\Delta
\end{equation}

\begin{figure}[h]
	\centering
	\includegraphics[width=\linewidth]{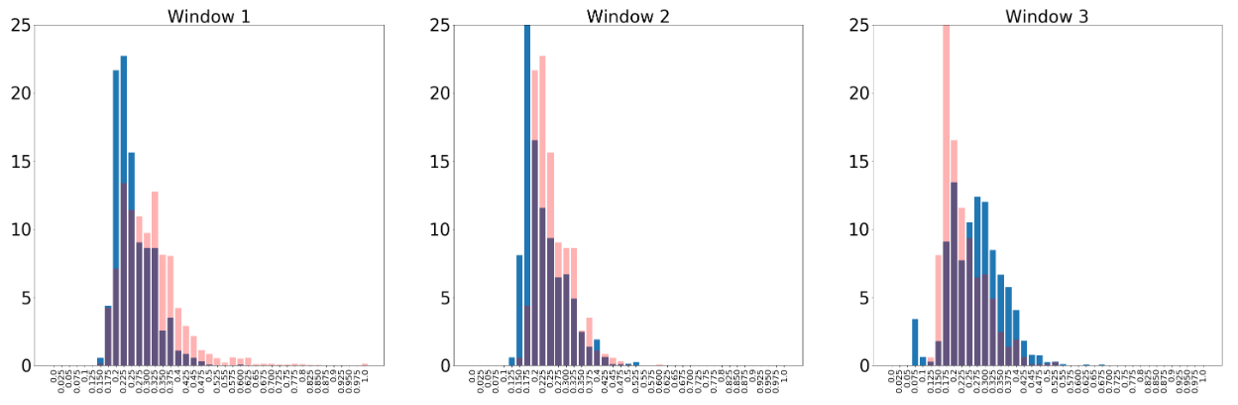}
	\caption{We show the difference in distance distribution ltiple windows. Each window is a set of 3000 data points that are relevant to disasters. The dark blue bars represent the window under consideration, while the pink bars are the prior window. In each case, there is enough divergence between points to constitute concept drift. Additionally, the distribution itself can be estimated using $\mathcal{N}(\mu,\sigma^2)$}
	\label{fig:rhobands}
\end{figure}

The intuition for using bands, as opposed to a spherical region is related to the curse of dimensionality in high dimensional data. Note that for some set of points in high-dimensional space, the volume of the unit hypersphere tends to zero\footnote{$V(d)=0.5^d\pi^{0.5d}/\Gamma(0.5d + 1)$}. So the majority of these points occur near the corners of the hypersphere. The $\Delta$-band then becomes a region around the centroid, where the hyperspherical region of radius $\delta_l$ (lower bound of $\Delta$) around the centroid is mostly empty. We approximate the $\Delta$-band of our data using $\mathcal{N}(\mu,\sigma^2)$, where $\mu, \sigma$ can be estimated based on the empirical observations in Figure~\ref{fig:rhobands}.

Then, with the $\Delta$-bands of two windows (current streaming window and classifier window), we can measure the relevancy of a classifier with the Kullback-Leibler (KL) divergence as follows. The standard KL metric is shown in Eq~\ref{eq:klmetric}, where the prior $P_A$ and posterior $P_B$ each model the data point $x_i$. The classifier window $w_C$ is the prior, and the current streaming window $w_S$ is the posterior.

\begin{equation}
D_{KL}(P_A||P_B)=-\sum_{x_i\in X}P_A(x_i)\log(P_B(x_i)/P_A(x_i))
\label{eq:klmetric}
\end{equation}

Then, let $x'_A=d(x_i,C_A)$, where $d$ is a distance metric and $C_A$ is the centroid of A. We obtain $x'_A$ and $x'_B$ from the prior and live distributions of $w_M$ and $w_S$, respectively, where each $x'$ is the distance of the data point from the centroids. We make the approximation $P_A(x')=\min(P(x'))$ if $P_A(x')=0$ to avoid KL discontinuity when $P_A(x')=0$. 

Each data point is a short-text string. We use word2vec to encode the string to $\Re^300$, and use Cosine Similarity as our distance metric since it is a more effective metric for word2vec. We can then use the divergence as a drift detector and classifier evaluator. We allow a smoothing period between windows to incorporate the new stream after drift detection. For each data point, we add it to the current window and update the window's $\Delta$-band. We then compare it to the window during the smoothing period to measure the divergence, where significant divergence indicates drift is occurring. 

In the case of drift, we create a new window and generate and update classifiers. In the absence of drift, we measure the distance between the data point and the centroids of all classifiers to obtain the top $k$ classifiers for dynamic ensemble creation.

\subsubsection{Classifier Generation and Update}
If drift is detected, we require changes to the classifiers that model the drifting data. We also generate new classifiers for the drifted data to avoid relying only on old classifiers that include drifted and old knowledge in their parameters. 

We can model any learning model or classifier $M$ as a mapping $f_M:\mathcal{M}\rightarrow\mathcal{Y}$ from the training and testing data ($\mathcal{M}$) to their respective class labels $\mathcal{Y}$. Here each $\mathcal{M}$ specifies a region in the data space. The traditional learning methods have characterized $\mathcal{M}$ as representative of the universe of data points (see \textbf{Related Work} section). This assumption is not suitable for the streaming data with noise and drift, where the training data distribution in one window may be different from the distribution in another window (see Figure
~\ref{fig:rhobands}). S any window contains only a subset of the data.\

We address this assumption without detection and generation approach, where we build a continuously evolving set of mappins, or classifiers, from the data space to labels. With drift detection, whenever the distribution of a region in the data space changes, we change the classifier associated with it. If a new distribution is discovered (i.e. new points do not belong in any $\Delta$-band in the mappings database), then we generate a new mapping for that region. We use the following algorithm for classifier generation and update.

\begin{algorithm}[h]
	\caption{Updating existing classifiers and generating new classifiers}
	\label{alg:virtualdrift}
	\begin{algorithmic}[1]
		\STATE $\mathtt{Parameters}$: $d$ (the distance metric, e.g. $\mathtt{CosineSimilarity}$); $k$-model selection policy $S_k$; $\lambda$
		\STATE $\mathtt{Inputs}$: $N$ Current models $\{M\}^N$, new data point $x_i$
		\STATE $\{M\}^k = S_k(x_i)$
		\STATE $mem\_xi = \mathtt{False}$
		\FOR{$M_j \in\{M\}^m$}
		\STATE \COMMENT{$D_{M_j}$ is the training data of model $M_j$, with $\Delta$-band $[\delta_l^j, \delta_h^j]$}
		\STATE $d'_{x_i} = d(x_i, D_C^j)$ \COMMENT{Distance to centroid}
		\STATE \COMMENT{Check if inside $\Delta$-band}
		\IF {$\delta_l^j < d'_{x_i} < \delta_h^j$}	
		\STATE $\mathcal{D}_{M_j} = \mathcal{D}_{M_j} \cup x_i$ \COMMENT{Add point to model's data}
		\STATE $\mathtt{Update(}M_j\mathtt{)}$ \COMMENT{Update classifier if $x_i$ is labeled by corroborative sources}
		\STATE $mem\_xi = \mathtt{True}$	\COMMENT{Flag to indicate data point has an associated region}
		\ENDIF
		\IF {$\delta_h^j \geq d'_{x_i} < \lambda$}
		\STATE $\mathcal{D}_{M_j} = \mathcal{D}_{M_j} \cup x_i$
		\ENDIF
		\ENDFOR
		\IF {\NOT $mem\_xi$}
		\STATE $D_G = D_G\cup x_i$
		\ENDIF		
	\end{algorithmic}
\end{algorithm}

Algorithm~\ref{alg:virtualdrift} covers model updates. The parameters are the distance metric $d$, a model selection policy $S_k$, and a generalization parameter $\lambda$. The model selection policy $S_k:x_i\rightarrow \{M\}^k$ selects the $k$-best models to classify $x_i$. Some examples of an ensemble selection policy include: set of all \textit{recent} models (where \textit{recent} indicates models created in the prior drift detection update step); high performing models over the entire set of models; high performing \textit{recent} models; $k$-nearest models based on distance between data point and centroids of the model's data window; or nearest $\Delta$-band models where only models whose $\Delta$-band contains $x_i$ are considered. 


For each point, we identify the $\Delta$-band the point belongs to (Lines 7-9). If an $x_i$ does not belong to a $\Delta$-band, we check if it belongs in a generalization band around the $\Delta$-band in Line 14, where we consider the region $[\delta_h^j,\lambda]$ just outside the $\Delta$-band. If an $x_i$ does not belong in any $\Delta$-band, we add it to the general memory $\mathcal{D}_G$ in Line 19, which we use to train new classifiers. The general memory are regions of the data space not yet seen in any existing model's data; it is used to create new classifiers when drift is detected using Eq~\ref{eq:klmetric}. When drift is detected, we address it for each model by using the data in its respective data (updated in Lines 10 and 15).  

\subsection{Implementation}
\begin{figure}[h]
	\centering
	\includegraphics[width=\linewidth]{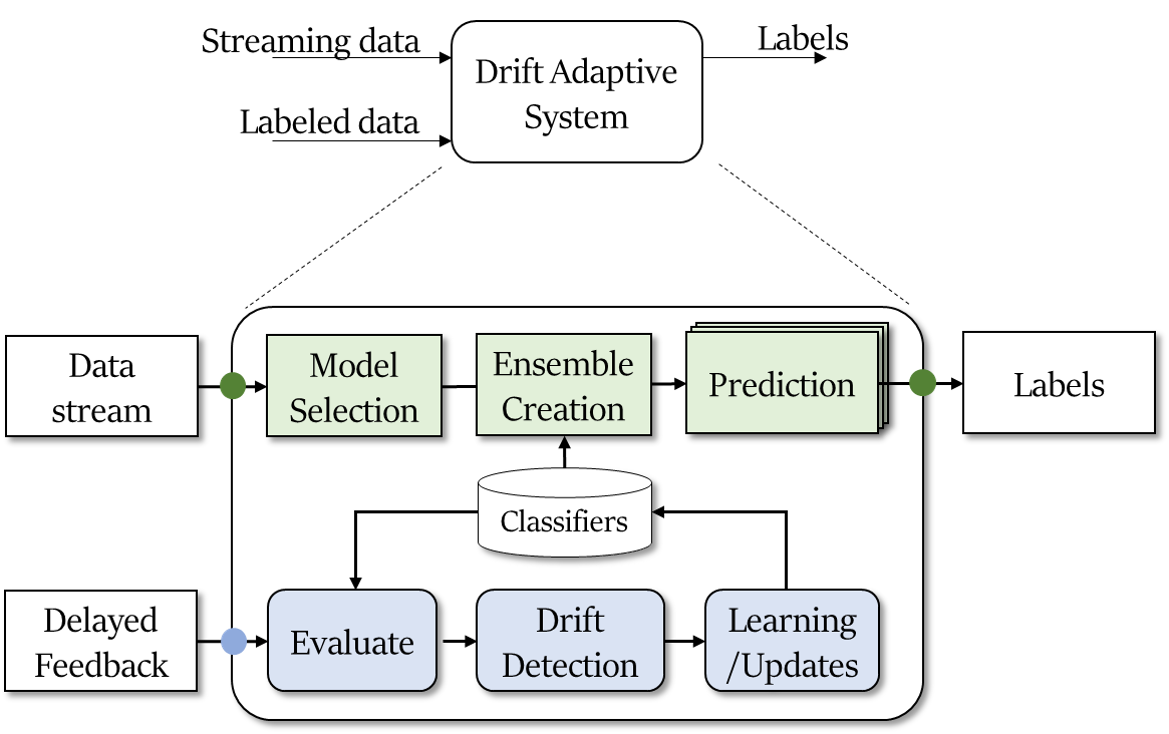}
	\caption{The drift adaptive system takes streaming data and labeled data, the latter being corroborative events obtained from corroborative sources. We use the corroborative events to evaluate existing classifier and fine-tune them continuously. The streaming data is used for real-time dense global prediction.}
	\label{fig:localsystem}
\end{figure}

We now describe the implementation of the teamed-classifier drift adaptive system, as described in Figure~\ref{fig:localsystem}. The drift adaptive system accepts two input streams - the real-time data stream (streaming data in Figure~\ref{fig:systemoverview}) and the delayed feedback labeled stream (corroborative events in Figure~\ref{fig:systemoverview}).

\paragraph{Real-time stream}
We use the classifier selection and drift detection approach described in the previous approach for the real-time stream to deliver predictions:

\begin{itemize}
	\item \textbf{Model selection}: We examine several model selection policies and use the $k$-nearest approach to select the models whose centroids are closest to a given data point.
	\item \textbf{Ensemble creation}: The $k$-models selected in the prior step are weighted on their performance $\omega$ on their datasets, multiplied by the distance: $w_k=\omega\cdot d(x_i,D_{M_k})$. The weights are normalized using the softmax function.
	\item \textbf{Prediction}: The dynamically generated ensemble's predictions are sent to the \textbf{Integrated Knowledgebase} in Figure~\ref{fig:systemoverview}. 
\end{itemize}

Simultaneously, we perform continuously perform classifier maintenance using delayed feedback from corroborative events (we call this delayed feedback since corroborative events are far slower than the real-time stream).

\begin{itemize}
	\item \textbf{Evaluate}: We can retroactively assign labels to real-time data points using corroborative events with the spatio-temporal assigmnent approach from Figure~\ref{fig:spatiotemporal}. We then use this label assignment to evaluate classifiers on performance.
	\item \textbf{Drift Detection}: Performance degradation entails explicit concept drift detection \cite{gama_drift_a}. In conjunction with the unsupervised drift detection with the KL metric in Eq~\ref{eq:klmetric}, we identify drift windows to generate new data memories (see Algorithm~\ref{alg:virtualdrift})
	\item \textbf{Learning/Updates}: Drift in existing data memories entails model update and new model generation on the corroboratively labeled data. New data regions (the general memory $\mathcal{D}_G$) discovered in the previous window are used exclusively for new mode creation, since they have no existing models to update.
\end{itemize}
	
\section{Evaluation}
We will first describe some drift charateristics of our data. Then we will cover further system implementation and neural network classifier details. We will briefly cover accuracy results on individual windows. Finally we will describe our end-to-end system, with a demo available at \textit{https://grait-dm.gatech.edu/demo-multi-source-integration/}.

\subsection{Drift Characteristics}
Since our physical event detection data is a raw real-time stream from social sensor sites, we face significant noise and drift in our data. If we rely on authoritative corroboration for the streaming data, we lose valuable time in event detection; as such, our system must be capable of adapting to such noise and drift continuously without human intervention. We have covered the approach in the prior section.

\begin{figure*}
	\begin{center}
		\includegraphics[width=.75\textwidth]{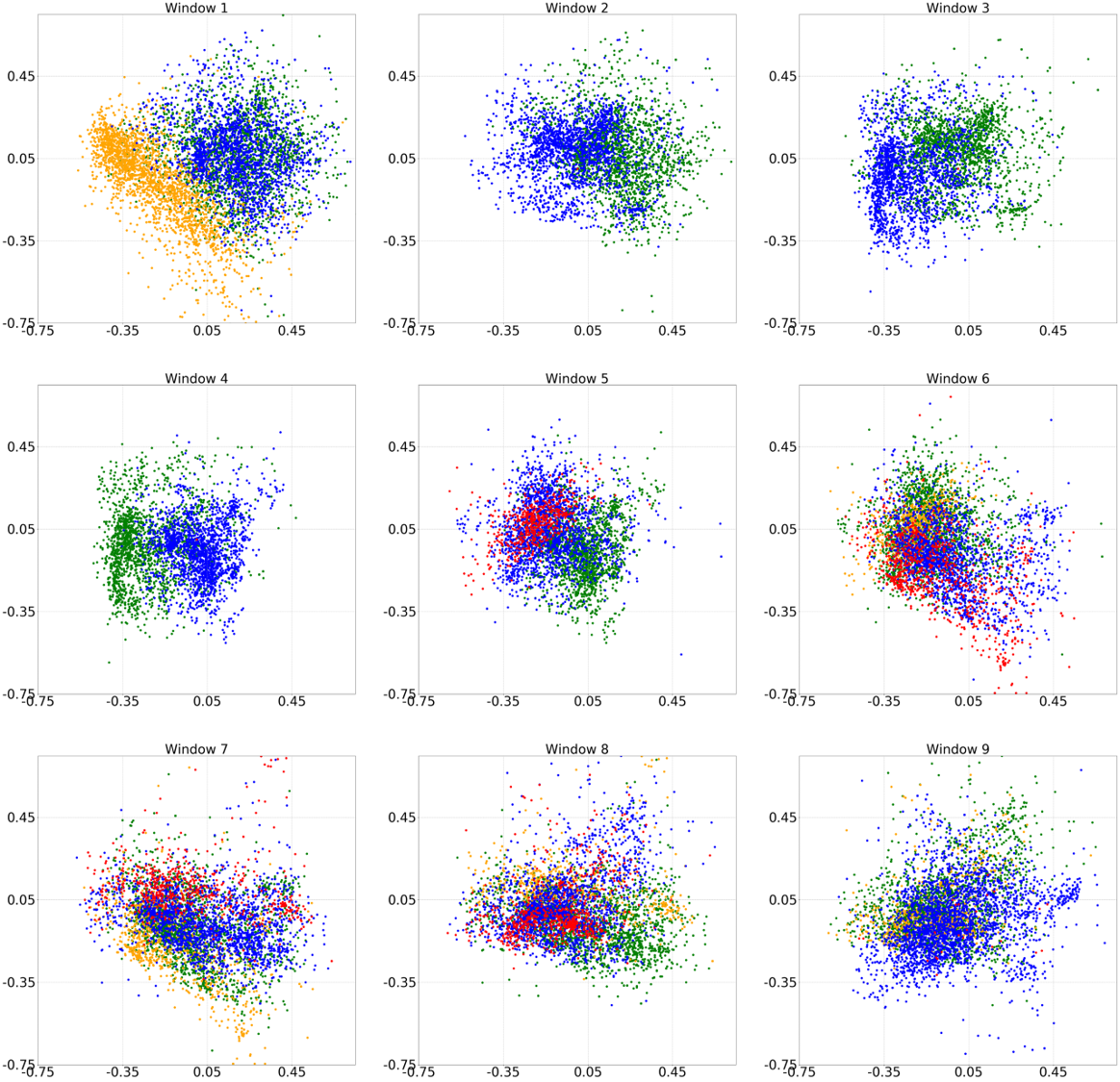}
	\end{center}
	\caption{Drift across multiple windows shown with t-SNE embedding. The axes represent raw componnt t-SNE scores and do not have semantic meaning other than distance between features.}
	\label{fig:acrosswindows}
\end{figure*}

We show the almost continuous drift for our data in Figure~\ref{fig:acrosswindows}. Each window is 3000 data points, and we show only a subset of windows. The red points are from the prior window, and the blue points are from the current window. For each window, we use t-SNE to embed both previous window and current window to 2D, and display them on the same plot. Some windows (Window 8, Window 9) do not show significant drift, with current window data occupying mostly the same space as previous window. However, other windows (Window 1, 4, 5, 6, 7) show more significant drift. Each point is a word2vec embedding of the associated string from social sensor. In each window image, blue points are positive samples in the current window, red points are negative samples in the current window, green points are positive samples in the previous window, and yellow points are negative samples in the previous window. We have noted in our paper the difficulty in automatically labeling negative samples, due to coverage considerations. As such, there is lower density of negative samples throughout compared to positive samples, creating a class imbalanced problem that adds to the existing drift and noise challenges.

\begin{figure}[h]
	\centering
	\includegraphics[width=\linewidth]{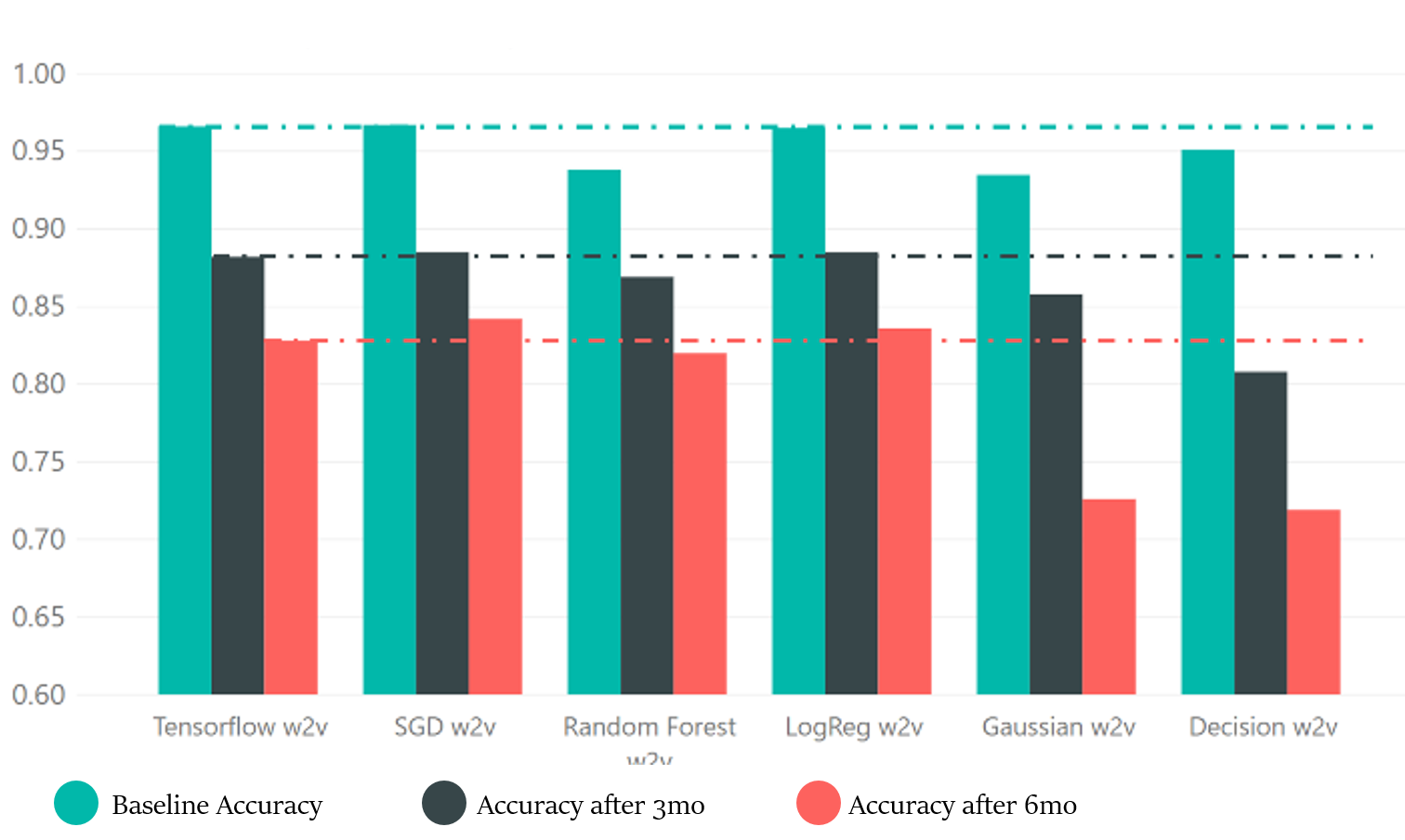}
	\caption{We test several classifiers outside our drift adaptive system. In each case, performance drops across a few months compared to the baseline accuracy. The Decision Tree, which relies on boundary conditions around features, suffers the most due to changes in feature distribution.}
	\label{fig:badperf}
\end{figure}

We also evaluate performance of a non-drift adaptive system on the drifting data, shown in Figure~\ref{fig:badperf}. We use a variety of linear classifiers such a SGD, Random Forest, Logistic Regression, Naive Bayes, and Decition Trees, along with neural networks. For linear classifiers, we perform grid search to obtain the best hyperparameters. We use an ensemble of text-classification networks \cite{fasttext,convclass,convclass2} for the neural network (using TensorFlow). As we note, each classifier suffers significant performance drops. 

We find this performance drop is due to a few factors: 
\begin{itemize}
	\item The social sensor data is noisy and has low context, yielding poor initial performance. As such, classifiers fail to generalize due to the variability in text streams from different regions or demographic groups (this diffusion is covered in part in \cite{lex_diff}).
	\item Heuristic or simple filtering rules are lacking; it is difficult to adapt heuristics to memetic changes (e.g. the word \textit{flood} and \textit{death} can be a good heuristic for the disaster event; however more recently, they have been used in conjunction with controversial political language, skewing the social sensor data)
	
\begin{table*}[]
	\centering
	\label{tab:perf}
	\caption{We show performance across multiple windows, and compare the baseline performance against our adaptive system. We find that in each window, our adaptive system excels against the baseline. We use the corroborative events to retroactively label points whev available, and use thse to perform classifier fine-tuning and updates as described in the prior section. We show that even with $<$5\% of labeled points, e are able to continuously improve performance from the static to the adaptive method.}
	\begin{tabular}{|l|rr|rr|rr|}
		\hline
		\multicolumn{1}{|c|}{\multirow{2}{*}{\textbf{Window}}} & \multicolumn{2}{c|}{\textbf{Performance}}                  & \multicolumn{2}{c|}{\textbf{Statistics}}                           & \multicolumn{2}{c|}{\textbf{Improvement}}                         \\
		\multicolumn{1}{|c|}{}                                 & \multicolumn{1}{c}{Static} & \multicolumn{1}{c|}{Adaptive} & \multicolumn{1}{c}{Unlabeled} & \multicolumn{1}{c|}{Corroborative} & \multicolumn{1}{c}{\% Labeled} & \multicolumn{1}{c|}{Improvement} \\ \hline
		Baseline                                               & 0.91                       & 0.97                          & NA                            & NA                                 & NA                             & NA                               \\
		1 Mo                                                   & 0.70                       & 0.88                          & 7205                          & 189                                & 2.62\%                         & 125.5\%                          \\
		2 Mo                                                   & 0.57                       & 0.90                          & 14245                         & 106                                & 0.74\%                         & 159.2\%                          \\
		3 Mo                                                   & 0.58                       & 0.90                          & 4867                          & 193                                & 3.97\%                         & 156.7\%                          \\
		4 Mo                                                   & 0.70                       & 0.88                          & 15847                         & 249                                & 1.57\%                         & 126.1\%                          \\
		5 Mo                                                   & 0.38                       & 0.86                          & 7084                          & 885                                & 12.49\%                        & 225.7\%                          \\
		6 Mo                                                   & 0.75                       & 0.99                          & 4873                          & 223                                & 4.58\%                         & 132.0\%                          \\ \hline
	\end{tabular}

\end{table*}

	\item The raw stream data covers millions of true physical events, where our desired class (disaster, specifically individual disasters such as flooding, landslides, wildfires) consists of a fraction of samples. Further, it is difficult to use trend analysis tools to perform detection since each instance of an event is a weak-signal event, with only 1-2 posts associated with it.
\end{itemize} 

\subsection{Performance}
We implement the end-to-end drift adaptive system described in Figure~\ref{fig:localsystem} and evaluate its performance across windows. The system as described integrates two streams: corroborative sources (i.e. news articles) and probabilistic supporting sources (i.e. predictions from ML classifiers) to deliver real-time predictions. Our system is not a retrospective trend analysis system such as the earthquake detector in \cite{dis_mgmt_sakaki}; rather, it is a continuously evolving, real-time system.

\begin{figure}[h]
	\centering
	\includegraphics[width=\linewidth]{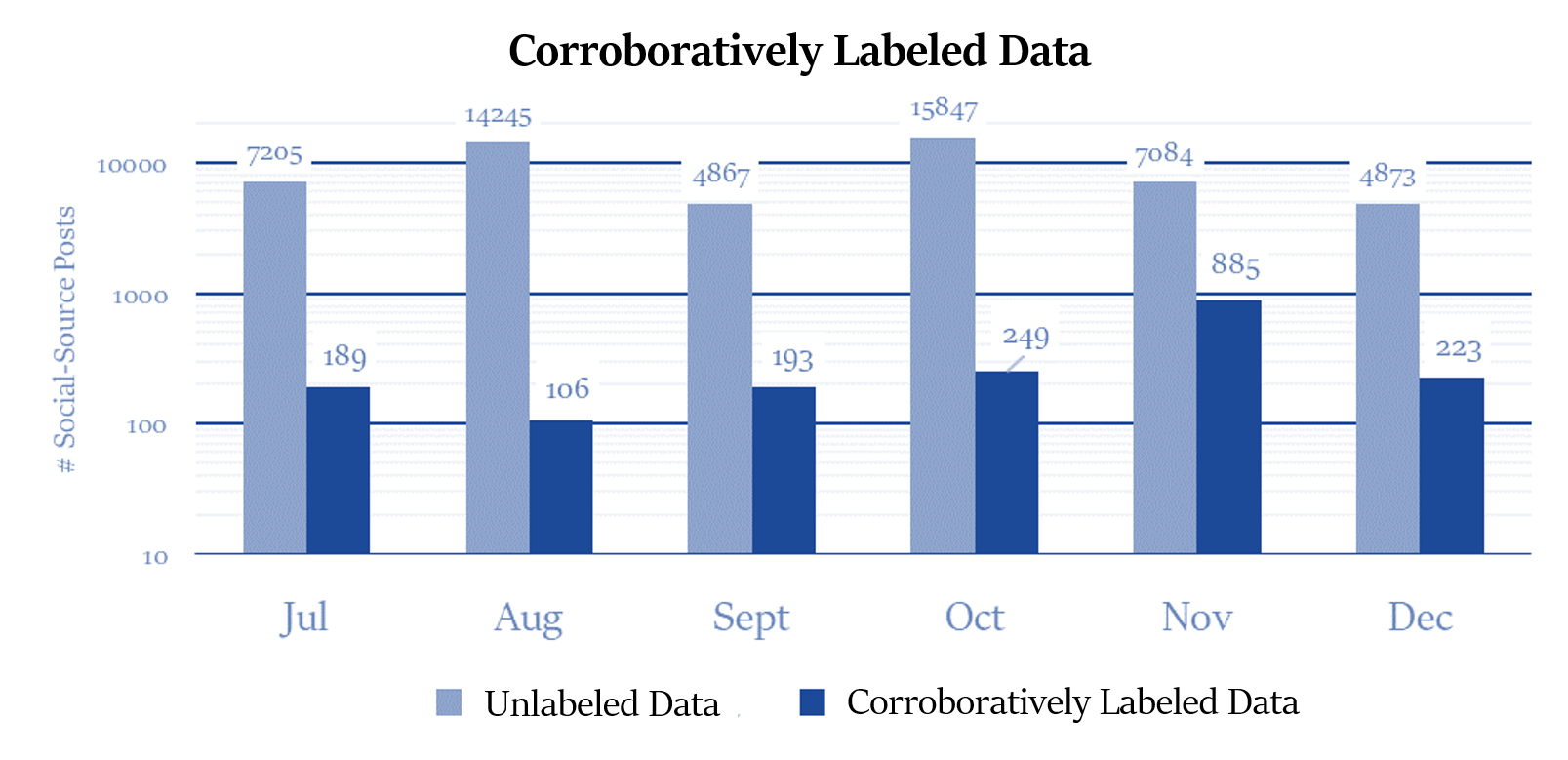}
	\caption{Only a fraction of points can be labeled with corroborative events (log-scaled y-axis).}
	\label{fig:corroborative}
\end{figure}

We have noted that corroborative events can be used to automatically label social sensor posts. We find that while this is true, they account for only a fraction of all social posts; their delay and lack of dense, global coverage prevents their use as a reliable oracle for labels. We show the difference between the raw stream and oracle-labeled points in Figure~\ref{fig:corroborative}, where often, less than 1\% of points could be so labeled. The remaining need to be processed with the drift adaptive system.

We show performance evaluation in Table 1, where it is clear our drift adaptive system exceeds the static performance. 

\subsection{End-to-end system}

\begin{figure}[h]
	\centering
	\includegraphics[width=\linewidth]{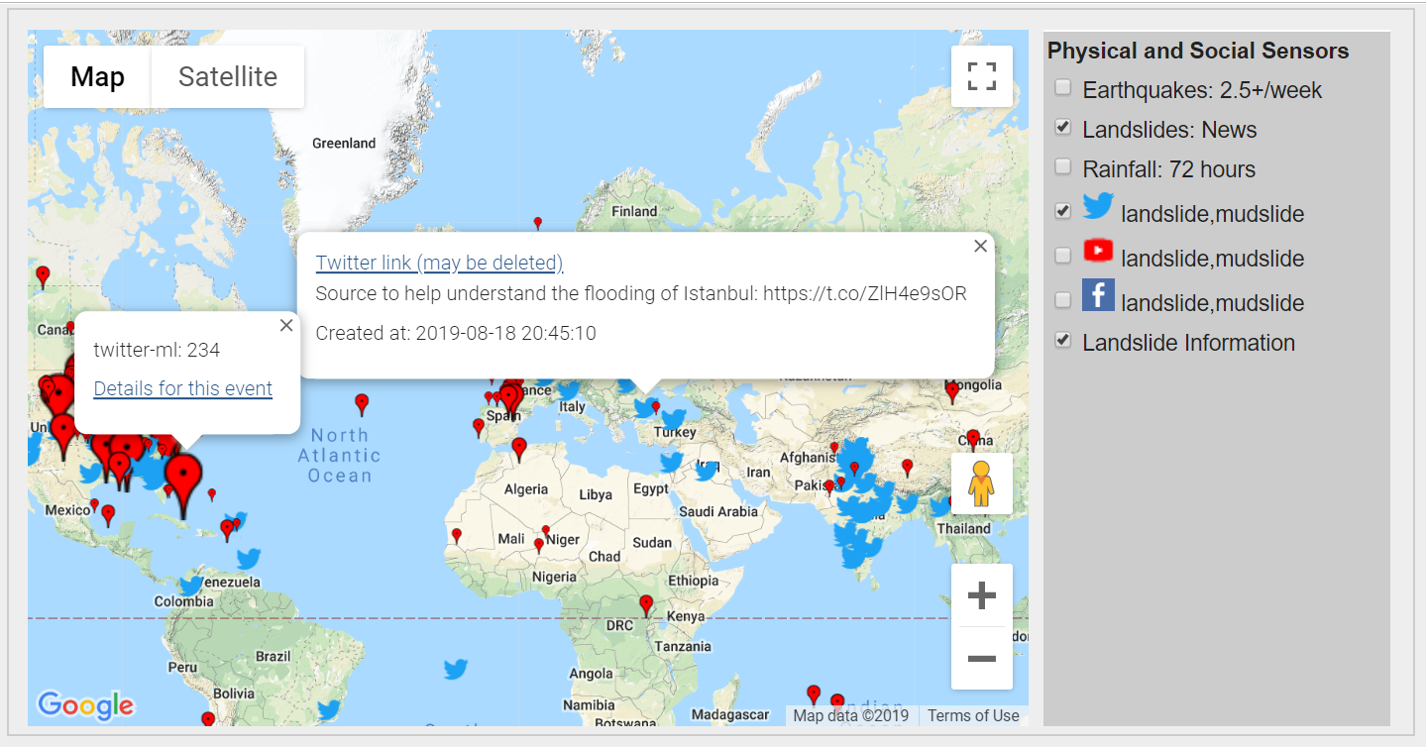}
	\caption{A screenshot of the LITMUS landslide detection system demo incorporating our drift adaptive system.}
	\label{fig:demo}
\end{figure}
We show in Figure~\ref{fig:demo} a screenshot of our end-to-end system available at \textit{https://grait-dm.gatech.edu/demo-multi-source-integration/}. Our system is resilient to drift, as we showed in Table~\ref{tab:perf}, and continues to function at high accuracy over six years after inception without any human intervention.

\begin{figure}[h]
	\centering
	\includegraphics[width=\linewidth]{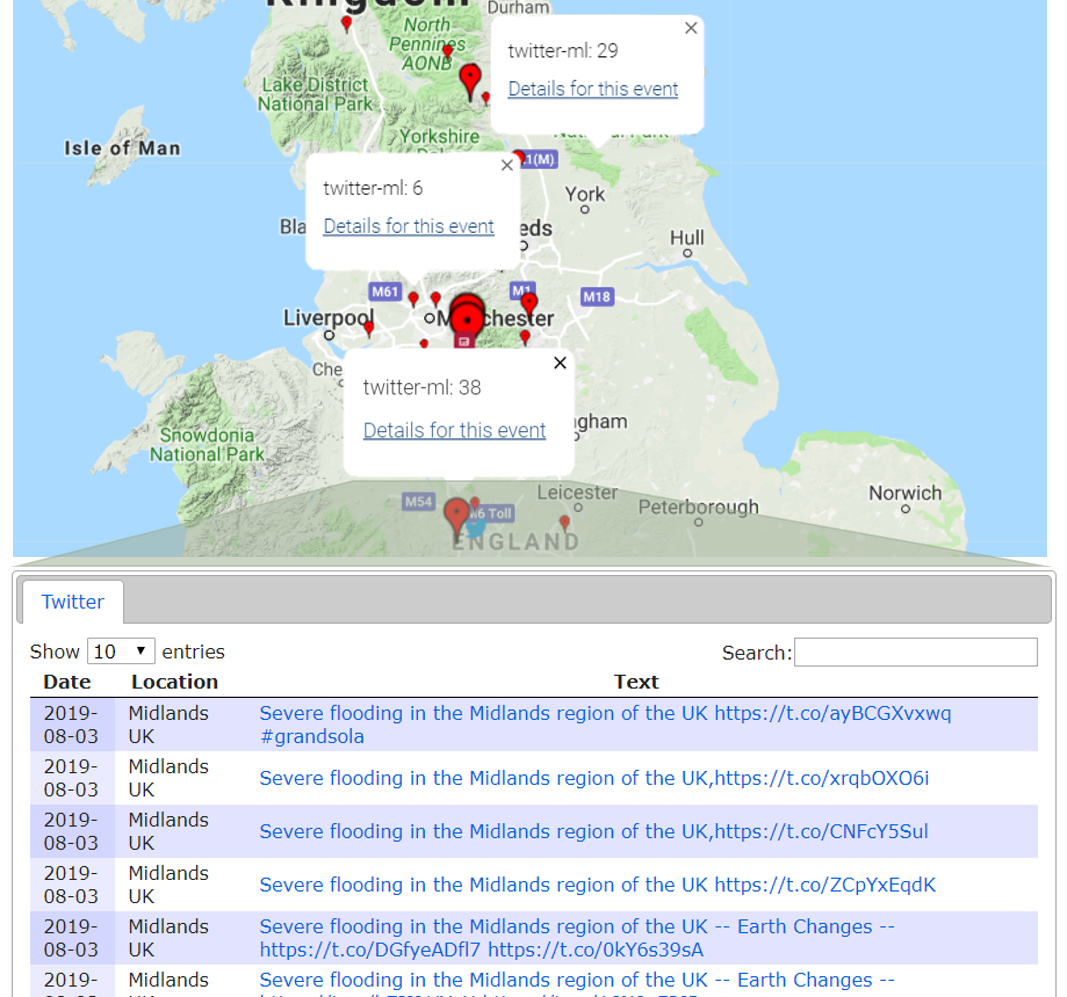}
	\caption{Flood event detection in the LITMUS platform. The highlighted event is detected only with social posts.}
	\label{fig:event}
\end{figure}

We also show an example of a detected flooding event in Figure~\ref{fig:event}, where the drift adaptive system has identified several flooding events in the UK. We also show below the map the collection of social posts that contributed to the event detection.

\section{Conclusions}
We have described an end-to-end drift-adaptive system for true physical event detection. Our approach integrates corroborative sources and probabilistic supporting sources to perform real-time physical event detection. Furthermore, our approach is able to adapt to the concept drift phenomena without sacrificing performance and without human labeling bottleneck as required in traditional drift adaptation techniques. We have implemented our system as a disaster detection application, with an online demo. Our approach does not make any limiting assumptions about its data, and performs detection in adversarial conditions where: (i) the data is noisy and drifting, (ii) the drift type is unknown and unbounded, (iii) feedback is limited and may not be available in most cases, and (iv) the events exhibit weak-signal characteristics. Our system is able to main high accuracy ($\sim$90\% f-score)  across multiple time windows without human intervention to perform fine-tuning or updates.

Our next steps include developing a management interface for our drift adaptive system to better examine system components and perform log analytics to improve real-time performance and scalable prediction delivery.


\bibliographystyle{IEEEtran}
\bibliography{main}

\end{document}